\documentclass[letterpaper]{article} 

\usepackage{misc}  
\usepackage{times}  
\usepackage{helvet}  
\usepackage{courier}  
\usepackage[hyphens]{url}  
\usepackage{graphicx, nicefrac} 
\usepackage[dvipsnames]{xcolor}
\usepackage{natbib}
\definecolor{ind_navyblue}{rgb}{0,0,0.50}
\definecolor{ind_green}{rgb}{0.07,0.53,0.31}

\usepackage{hyperref}

\hypersetup{colorlinks,breaklinks,
         urlcolor=ind_navyblue,
         linkcolor=ind_navyblue,
         citecolor=ind_navyblue}
         
\usepackage{caption} 
\usepackage{booktabs}
\usepackage{helvet}

\usepackage{algorithm}
\usepackage{algorithmic}
\usepackage{amsfonts}
\usepackage{newfloat}
\usepackage{listings}
\DeclareCaptionStyle{ruled}{labelfont=normalfont,labelsep=colon,strut=off} 
\floatstyle{ruled}
\newfloat{listing}{tb}{lst}{}
\floatname{listing}{Listing}
\title{Policy composition in reinforcement learning \\ via multi-objective policy optimization}
\author{
    Shruti Mishra\thanks{Work done during an internship at DeepMind, now at Sony AI. \href{mailto:shruti.mishra10@alumni.imperial.ac.uk}{shruti.mishra10@alumni.imperial.ac.uk}},
    Ankit Anand,
    Jordan Hoffmann\thanks{Work done while at DeepMind.}, \\
    Nicolas Heess, 
    Martin Riedmiller,
    Abbas Abdolmaleki,
    Doina Precup
}
\affiliations{
    Google DeepMind
}

\begin{document}
\renewcommand{\sectionautorefname}{Section}
\renewcommand{\subsectionautorefname}{Section}
\renewcommand{\subsubsectionautorefname}{Section}
\newcommand{\algorithmautorefname}{Algorithm}

\maketitle

\begin{abstract}
    We enable reinforcement learning agents to learn successful behavior policies by utilizing relevant pre-existing teacher policies. The teacher policies are introduced as objectives, in addition to the task objective, in a multi-objective policy optimization setting. Using the Multi-Objective Maximum a Posteriori Policy Optimization algorithm \citep{abdolmaleki2020distributional}, we show that teacher policies can help speed up learning, particularly in the absence of shaping rewards. In two domains with continuous observation and action spaces, our agents successfully compose teacher policies in sequence and in parallel, and are also able to further extend the policies of the teachers in order to solve the task.
        
    Depending on the specified combination of task and teacher(s), teacher(s) may naturally act to limit the final performance of an agent. The extent to which agents are required to adhere to teacher policies are determined by hyperparameters which determine both the effect of teachers on learning speed and the eventual performance of the agent on the task. In the {\tt humanoid} domain \citep{deepmindcontrolsuite2018}, we also equip agents with the ability to control the selection of teachers. With this ability, agents are able to meaningfully compose from the teacher policies to achieve a superior task reward on the {\tt walk} task than in cases without access to the teacher policies. We show the resemblance of composed task policies with the corresponding teacher policies through \href{https://www.tinyurl.com/anonrl}{videos}.
\end{abstract}

\section{Introduction}
In recent years, reinforcement learning (RL) agents have been trained to perform tasks to a level that meets or exceeds the abilities of human experts, particularly in the context of games \cite{wurman2022outracing,vinyals2019grandmaster,silver2018general,silver2017mastering,silver2016mastering,mnih2015human}.
In simulated robotics and animal-like domains, RL agents have been trained from scratch to move through their respective environments using various types of locomotion and navigation behaviors, involving crawling \cite{mishra2020coordinated}, gliding \cite{reddy2016learning}, running \cite{barth2018distributed,schulman2017proximal}, swimming \cite{barth2018distributed,schulman2017proximal,colabrese2017flow,verma2018efficient}, and traversing uneven terrain \cite{heess2017emergence}.
RL agents in these domains are characterized by significant simplifications in the mathematical models of the physics engines that create the environment, e.g., \cite{todorov2012mujoco}, resulting in unnatural behaviors \cite{barth2018distributed,schulman2017proximal,heess2017emergence}, or by simplified observation and action spaces, specific to the task in consideration, rather than the agent and environment in general \cite{colabrese2017flow,giardina2021models,mishra2020coordinated,novati2019controlled,reddy2016learning}. 

Limitations to the environmental complexity that RL agents can cope with are, in part, a result of the large amount of experience required by RL agents before they are able to execute tasks adequately. 
As such, efforts to improve sample efficiency in RL algorithms hold the promise to enable agents that can execute tasks in more realistic sensorimotor settings, both for understanding animals, and understanding and creating robots. Moreover, in biological and biologically-inspired domains, the learning of tasks can rely on policies that are successfully able to solve related tasks -- a tiger cub learning to hunt does not have to relearn how to walk and run, but can call upon those behaviors it is already able to execute. Similarly, we might imagine that a robot that is learning to solve a navigational task does not need to relearn how to locomote in its specified domain. As noted in prior work, e.g.~\citet{barreto2020fast, peng2019mcp, qureshi2020composing}, it makes sense for RL agents to reuse knowledge to make learning more efficient.

Towards leveraging existing knowledge for learning tasks using RL, we equip agents with some access to pre-existing policies for related tasks in the same environment. We formulate the problem of leveraging pre-existing policies using multi-objective policy optimization where an additional objective is introduced corresponding to adherence to each pre-existing policy. The additional objectives are formalized as the KL divergence between the agent policy and the corresponding pre-existing policy. The hyperparameters for each KL divergence term (corresponding to each teacher) control the adherence to the relevant pre-existing policy. Pre-existing policies can be leveraged sequentially or concurrently. We empirically demonstrate the usefulness of this formulation on point mass and humanoid domain in the DeepMind control suite~\cite{deepmindcontrolsuite2018}. Additionally, adhering too closely to pre-existing policies may limit the performance of an agent in a new task. To overcome this, in the humanoid domain, we equip the agent with the ability to control the selection of teachers by learning the weight of KL divergence (between agent's policy and each pre-existing policy) for each pre-existing policy. We observe that the agent where the adherence to pre-existing policies is learned using the task objective not only achieves superior performance to an agent without access to pre-existing policies, but also able to match the performance of an agent where the adherence to pre-existing policies is handcrafted. Overall, a multi-objective framework provides a flexible way to compose teacher policies in contrast to existing work, e.g.~\citet{qureshi2020composing, peng2019mcp}, that uses pre-existing policies as primitives. Moreover, the framework is general enough to not only adhere to teachers but also disincentivize certain configurations by through a negative weight of the KL divergence between the agent's policy and a pre-existing policy. We make following contributions in this work:
\begin{enumerate}
    \item We formulate policy composition using pre-existing policies as multi-objective optimization problem using the MO-MPO algorithm of \citet{abdolmaleki2020distributional}. 
    \item We illustrate the usefulness of the above formulation in composing pre-existing policies both sequentially and concurrently in domains from the DeepMind control suite~\cite{deepmindcontrolsuite2018} and show impressive gains in sample efficiency.
    \item We demonstrate how an adherence to pre-existing policies can be learned by making the weight of KL divergence between the agent policy and corresponding pre-existing policy a function of observation. Our experiments demonstrate comparable empirical performance to handcrafted selection of these weights.
    \item We show agents that are able to compose between discontinuous policies, thus demonstrating flexibility over approaches that consider pre-existing policies as a primitive layer to be built upon. 
\end{enumerate}

\section{Method}
\label{sec:comp-method}
Motivated by considerations of computational efficiency and biological plausibility, we seek to develop a method for incorporating pre-existing \emph{skills} flexibly in solving a \emph{task} using RL.

\subsection{Notation}
\label{sec:method-notation}
A task is described using the framework of a Markov Decision Process (MDP) \cite{puterman2014markov}, which is defined as a tuple,
\begin{equation}
    \mathcal{M} \equiv \left(\mathcal{S}, \mathcal{A}, p, r, \gamma \right), 
\end{equation}
where $\mathcal{S}$ and $\mathcal{A}$ are the state and action spaces, respectively, of an agent in an environment, $p(s'|s,a)$ and $r(s,a) \in \mathbb{R}$ specify the probability distribution of the next state $s'$ and the expected reward, respectively, arising from taking an action $a$ in a state $s$. For a specified MDP, an agent's objective is to learn a policy $\pi(a|s)$ that maximizes a discounted sum of rewards,
\begin{equation}
    \sum_{t=0}^{T} \gamma^t {\mathbb{E}}_{\pi(a|s)} \left[r_t\right],
    \label{eq:comp-obj}
\end{equation} 
with $\gamma$ being the discount factor, and $t$ denoting the index of the discrete steps taken by the agent in the space of states and actions. 

We are concerned with scenarios where an RL agent has access to pre-existing skills, in the form of policies. We refer to a pre-existing policy function as a \emph{teacher policy}, $\pi_{\mathrm{teacher},i}(a|o)$, for taking an action $a \in \mathcal{A}$ for a given observation $o \in \mathcal{O}$, where $\mathcal{A}$ and $\mathcal{O}$ are the action and observation spaces, respectively. Although, observation is output of sensors and state is the complete world state information, in this work, for simplicity and the domain under consideration, we use both terms interchangeably for the rest of paper.
Transitions take place from one state to the next, and the policy is executed over the space of observations. 
An RL agent does not, a priori, have a policy that can solve the task. It may use the teacher policies, $\pi_{\mathrm{teacher},i}(a|o)$, to inform its learning.

We consider tasks in continuous-control domains, where the observation and action spaces are both vectors with components that are real-valued, i.e., $\left(\mathcal{O}, \mathcal{A}\right) \in \left(\mathbb{R}^{|\mathcal{O}|}, \mathbb{R}^{|\mathcal{A}|}\right)$, with $|\mathcal{O}|$ and $|\mathcal{A}|$ being the number of components in the observation and action vectors, respectively. 

\subsection{Choice of algorithm}
Considering the problem of policy composition via multi-objective RL has the potential to have additional flexibility in composition in comparison to approaches that view pre-existing policies as a primitive layer to be built upon e.g.~\citet{qureshi2020composing,peng2019mcp}.
We use the Multi-Objective Maximum a Posteriori Policy Optimization (MO-MPO) algorithm \cite{abdolmaleki2020distributional}, a multi-objective actor-critic algorithm, to compose skills. Using this algorithm, the first objective in the MO-MPO framework is the standard RL task objective, as specified in \autoref{eq:comp-obj}. The subsequent objectives are the KL divergences from one or a number of teacher policies,
    \begin{equation}
    \label{eq:mompo-teacher}
         D_{\mathrm{KL}} \left(\pi(a|o)\|\pi_{\mathrm{teacher},i}(a|o)\right).
    \end{equation}

\begin{algorithm}[h]
\small
\caption{MO-MPO with teacher policies: One policy improvement step for the MO-MPO algorithm by \citet{abdolmaleki2020distributional} is shown below. Our specific implementations of each of the objectives in the context of the MO-MPO algorithm and the optional agent-selection of $\epsilon_{\mathrm{teacher},i}$ are indicated in \emph{italicized text}.}\label{alg:mo-mpo}
\begin{algorithmic}[1]
\STATE {\bf given} batch-size, L, number of actions, M, number of objectives, $N$, previous policy, $\pi_{\mathrm{old}}$, Q-functions, $\{{Q}_i^{\pi_{\mathrm{old}}}(o,a)\}_{i=1}^N$ preferences, $\{\epsilon_i(o)\}_{i=1}^N$, previous temperatures $\{\eta\}_{i=1}^N$, replay buffer, $\mathcal{D}$, first-order gradient-based optimizer, $Opt$,
\STATE {\bf initialize $\pi_\theta$ from the parameters of $\pi_{\mathrm{old}}$} 
\REPEAT
\item[]
\STATE {\bf // Step 1: Sample based policy (weights)}
\STATE {// Collect dataset $\{o^k, a^{kj}, {Q_i^{kj}}_{k, j, i}\}^{L, M, N}$, where}
\STATE // $M$ actions $a^{kj} \sim \pi_\mathrm{old}(a|o^k)$, $Q_i^{kj} = Q_i^{\pi_\mathrm{old}}(o^k, a^{kj})$ \emph{and optionally $\epsilon_i^{k} \sim \pi_\mathrm{old}(\epsilon_i|o^k)$ according to \autoref{fig:schematic_choose-teacher-epsilon}}
\item[]
\STATE {// Compute action distribution for each \emph{objective}}
\FOR{i = 1,...,$N$}
{
\IF{\emph{i = 1}}
\STATE \emph{Objective: $\sum_{t=0}^{T} \gamma^t {\mathbb{E}}_{\pi(a|o)} \left[r_t\right]$}
\ELSE
\STATE \emph{Objective: $D_{\mathrm{KL}} \left(\pi(a|o)\|\pi_{\mathrm{teacher},i}(a|o\right)$}
\ENDIF
}
\STATE $\delta_{\eta_i} \leftarrow
{\nabla_{\eta_i}}\eta_i \epsilon_i
+ \eta_i \sum_k^L\frac{1}{L} \left(
\sum_j^M \frac{1}{M} \exp\left(\nicefrac{Q_i^{kj}}{\mathrm{\eta_i}}\right)
\right)
$
\STATE Update $\eta_i$ based on $\delta_{\eta_i}$, using optimizer $Opt$
\STATE $q_i^{kj} \propto \exp\left(\nicefrac{Q_i^{kj}}{\mathrm{\eta_i}}\right)$
\ENDFOR
\item[]
\STATE {\bf // Step 2: Update parametric policy}
\STATE{
$\delta_\pi \leftarrow
-\nabla_\theta \sum_k^L \sum_j^M \sum_i^N q_i^{kj} \log \pi_\theta\left(a^{kj}|o^i\right)
$
}
\STATE {(subject to additional (KL) regularization)}
\STATE Update $\pi_\theta$ based on $\delta_\pi$, using optimizer $Opt$
\item[]
\UNTIL{fixed number of steps}
\STATE return $\pi_{\mathrm{old}}= \pi_\theta$
\end{algorithmic}
\end{algorithm}

According to the MO-MPO framework, the preference for the first objective, the task objective, is specified by the parameter $\epsilon$. Each of the subsequent objectives, denoted by index $i$, has a preference denoted by a parameter $\epsilon_i(o)$. The relative values of $\epsilon$  and $\epsilon_i$ specify the relative strength to which the MO-MPO agent is required to satisfy each of the objectives. The MO-MPO algorithm \cite{abdolmaleki2020distributional}, together with our specification of each objective, is summarized in \autoref{alg:mo-mpo}. The reader is referred to the work of \citet{abdolmaleki2020distributional} for further details on the algorithm. In the subsequent sections, we describe how the $\pi_{\mathrm{teacher},i}(a|o)$ are obtained and $\epsilon_i(o)$ are specified or learned.
\subsection{Teacher policies} 
Teacher policies are obtained by training an agent on an appropriately specified task, denoted to be the \emph{teacher task}. After the agent has learned a successful policy for the teacher task, this is frozen and used in \autoref{eq:mompo-teacher}. 
In general, teacher policies are limited in scope; imitating the teacher policy may not be sufficient to solve the task. Additionally, teacher policies may compete with each other and be counterproductive to solving the task in different parts of $\mathcal{O} \times \mathcal{A}$. 
For simplicity, our work is limited to the case where, for a particular task, the observation and action spaces, $\left(\mathcal{O}, \mathcal{A}\right)$, for the teachers are the same as those of the corresponding task. 

\subsection{Types of composition}
\label{sec:comp-method-comptypes}
We consider the scenarios where teacher policies are relevant to distinct parts of the observation space, and scenarios where teacher policies are relevant to overlapping parts of agent's observation space. In the framework of an MDP for continuous-control settings, these two scenarios approximate the temporal and spatial composition of policies, respectively. 
As such, our approach offers the flexibility of spatio-temporal composition.
The values of $\epsilon_i(o)$ in \autoref{eq:mompo-teacher} can be a choice made by the agent, or handcrafted as part of the learning algorithm. 
\paragraph{Agent control over the influence of teachers}
\begin{figure}[H]
    \centering
    \includegraphics[scale=0.9]{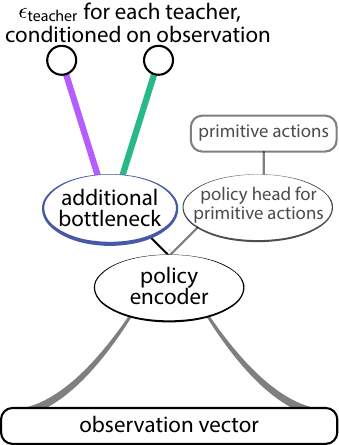}
    \caption{\textbf{Schematic for agent choosing the relevance of each teacher.} Observations, specified by a vector $o$, go through a neural network, the policy encoder. The output of the policy encoder, an observation embedding, is passed through $(i)$ policy head for the primitive actions $a$, and $(ii)$ an additional bottleneck, which finally outputs ${\epsilon}_{\mathrm{teacher},i}(o)$ for each teacher denoted by the index $i$. The distinct preferences for each teacher are indicated by green and purple colors, representing distinct preferences for the respective teacher policies. Parameters for the policy encoder and additional bottleneck are specified in Appendix A. 
    }
    \label{fig:schematic_choose-teacher-epsilon}
\end{figure}
In our work, an agent may control the influence of the teacher policy through a modification of the action space, $\mathcal{A}$, of the original MDP to include additional action components, $a_i =  \epsilon_{\mathrm{teacher},i}(o) \ \forall \ i \in \{0, \dots, {N_{\mathrm{teacher}}-1}\}$ , where $N_{\mathrm{teacher}}$ is the number of teachers. 
In controlling $\epsilon_{\mathrm{teacher},i}(o)$, the agent may encounter the following types of degeneracy: 
\begin{itemize}
    \item A distinct value of $\epsilon_{\mathrm{teacher},i}(o)$ is chosen for each observation. Allowing the agent to have this level of control is likely to lead to overfitting, since we start with the assumption that relevant teacher policies help shape learning for finite portions of the observation space. Additionally, if teacher policies are used for small or infinitesimal portions of the observation space, then the teacher(s) will not provide a useful signal. This is because requiring a unique action component, $a_i = \epsilon_{\mathrm{teacher},i}(o)$, for each observation, $o$, will essentially amount to learning a policy for the original MDP. 
    \item A value of $\epsilon_{\mathrm{teacher},i}(o)=0$. We allow the agent to have the flexibility to choose this, just as it has the ability to choose $a(o)=0$ for any other component of the action. 
\end{itemize}
To mitigate the first type of behavior, we induce a bottleneck from the observation space $\mathcal{O}$ to the space of parameters over which $\epsilon_{\mathrm{teacher},i}\left(o\right)$ can be chosen. A schematic of agent selection of $\epsilon_{\mathrm{teacher},i}(o)$ is shown in \autoref{fig:schematic_choose-teacher-epsilon}.
Furthermore, we require $\epsilon_{\mathrm{teacher},i}\left(o\right) \ge 0 \ \forall \ i,o$. This relies on the heuristic that the teacher policies $\pi_{\mathrm{teacher},i}$ are peaked around a small portion of the action space $\mathcal{A}$. As such, $\epsilon_{\mathrm{teacher},i}\left(o\right)<0$ means a penalization of agent policies away from a peaked probability distribution. In simulated domains, most environments are under-specified, meaning that a task can be solved successfully via several distinct policies. In such a setting, a penalty guiding an agent away from a specified policy is not a meaningful constraint. We note that in environments that do not share this feature, allowing $\epsilon_{\mathrm{teacher},i}\left(o\right) < 0$ may serve a meaningful purpose.

\section{Literature on skill composition}
\label{sec:comp-lit}
\begin{table*}[h]
    \center
    \caption{{\bf Q-values for 1-step MDPs} under primitive, scaled, and composed reward functions, $r$. For each reward, the action, $a$, with the highest value, $Q(a)$, is highlighted in {\bf bold text}.}
    \begin{tabular}{rccc} 
    \toprule
            & $Q\left(a^{(1)}\right)$ & $Q\left(a^{(2)}\right)$ & $Q\left(a^{(3)}\right)$ \\ \cmidrule{2-4}
        $r_1$ & \bf{0.6} & 0.4 & 0.0 \\
        $r_2$ & 0.0 & 0.4 & \bf{0.6} \\ \cmidrule{2-4}
        $r_{{2}, \ \mathrm{scaled}} = 10r_{2}$ & 0.0 & 4.0 & \bf{6.0} \\ 
        $r_{\mathrm{new}} = 0.5\left(r_{1} + r_{2}\right)$ & 0.3 & \bf{0.4} & 0.3\\
        $r_{\mathrm{new}, \ \mathrm{scaled}} = 0.5\left(r_{1} + r_{{2}, \ \mathrm{scaled}}\right)$ & 0.3 & 2.2 & \bf{3.0} \\ \bottomrule
    \end{tabular}
    \label{tbl:comp-bandit}
\end{table*}

As with most scientific endeavors, our work bears connections and parallels to work in many fields. While we touch on these in other sections, this section considers methods in the literature on composing skills using RL.
Broadly, there are two sets of approaches to composing skills: policy--based composition, e.g. \citet{qureshi2020composing,peng2019mcp}, and value--based composition, e.g., \citet{barreto2019option}. In both of these, the composition of skills is done by attributing weights to the underlying skills, and combining the weighted skills in an additive \cite{qureshi2020composing,barreto2019option} or multiplicative \cite{peng2019mcp} manner. The goal of the RL agent is to learn a combination of weights through the task objective, specified in terms of a reward $r$. 

The difference between policy--based methods and value--based methods arises in the notion of \emph{skill}, and how these may be combined. In policy--based methods, a skill $i$ is specified in terms of a policy $\pi_i(a|o)$. While, in general, the policy is stochastic, it does not change as a function of the task. In value--based methods, skills are combined using an internal notion of \emph{value}. This difference can be illustrated through a 1-step MDP, i.e., $K=0$ in \autoref{eq:comp-obj}. The value function $Q(s_0,a) \equiv r(a)$ for a single initial state $s_0$. We define reward functions $r^{(T_1)}=r_1$ and $r^{(T_2)}=r_2$ for two tasks, $T_1$ and $T_2$, in a discrete action space $\mathcal{A} \equiv \{a^{(1)}, a^{(2)}, a^{(3)}\}$. \autoref{tbl:comp-bandit} shows these example MDPs, as well as MDPs constructed by linearly combining the rewards associated with these primitive MDPs. In the combined MDPs, policy--based methods will choose from the actions corresponding to the primitive skills, in this case actions $a^{(1)}$ and $a^{(3)}$, whereas value--based methods will choose from actions corresponding to the highest value for the combined task reward, which also allows for the possibility of choosing action $a^{(2)}$. Moreover, value--based methods are sensitive to the scale of the rewards, which is shown by comparing the actions with the highest value in the last two rows of \autoref{tbl:comp-bandit}, which are different as the scale of the reward for $T_2$ changes. %

By defining a \emph{skill} to be a pre-existing teacher policy, $\pi_{\mathrm{teacher},i}(a|o)$, as in 
\autoref{sec:method-notation}, that can be accessed by an agent, our work falls under the approach of policy--based composition.

\section{Experiments} \label{sec:comp-exp}
In this section, we describe experiments conducted using the method specified in \autoref{sec:comp-method} and implemented using the Acme framework \cite{hoffman2020acme}. Each of our experiments is designed to highlight a particular type of composition, as categorised in \autoref{sec:comp-method-comptypes}. For set of experiments described here, we first specify the teachers used by the agent(s), then describe the experiments conducted using these teachers, followed by a description of the results.

\subsection{Domains and tasks}
We use two continuous-control domains from the DeepMind Control Suite \cite{deepmindcontrolsuite2018}. The domains and corresponding tasks are chosen to highlight a particular type of policy composition that can be achieved with our method.
The first domain is the {\tt{humanoid}}. We consider two tasks in this domain, {\tt stand} and {\tt walk}. For each of these tasks, the {\tt{humanoid}} is initialized in a pose of random joint configurations, some distance above the ground. It then falls to the ground under gravity and the agent must either act to obtain an upright orientation ({\tt stand} task) or maintain an upright orientation at a fixed value of forward velocity in its local frame of reference ({\tt walk} task).
The second domain is the {\tt point\_mass}, where a mass is initialized at a random location in a two-dimensional square arena. The agent's task is to minimize the distance of the {\tt point\_mass} from a specified target location. 

\subsection{Humanoid domain, using a stand teacher}
\label{sec:comp-exp-human-std}
In these experiments, we consider scenarios where teachers are relevant for part of the observation space, but may be uninformative, or counterproductive, to attaining the task objective in a different part of the observation space. This shows the capability of agents to use the task reward to go beyond what is specified via the available teacher policies. 

We use a single {\tt stand} teacher, which is obtained by training a MO-MPO agent on the {\tt humanoid} {\tt stand} task. 
For the {\tt stand} task, a {\tt stand} teacher provides a successful policy. For the {\tt walk} task, a {\tt stand} teacher provides a portion of a successful policy -- an agent that can get up after it has fallen on the ground. It must then further learn how to move forward at the relevant speed that corresponds to the {\tt walk} task. We ran experiments to see the effect of using a single {\tt stand} teacher policy on tasks in the {\tt humanoid} domain.  

To understand the effect of teachers on the learning, we modified the task reward to be sparse along the parameters that specify the upright orientation and head height of the {\tt humanoid}.
Specifically, setting {\tt upright\_margin = 0} and {\tt stand\_margin = 0} make the shaping reward sparse for the aspects of the task on maintaining an upright orientation of the torso and maintaining a head height above a threshold, respectively.

\begin{figure}[h!]
    \center
    \includegraphics[width=1\columnwidth]{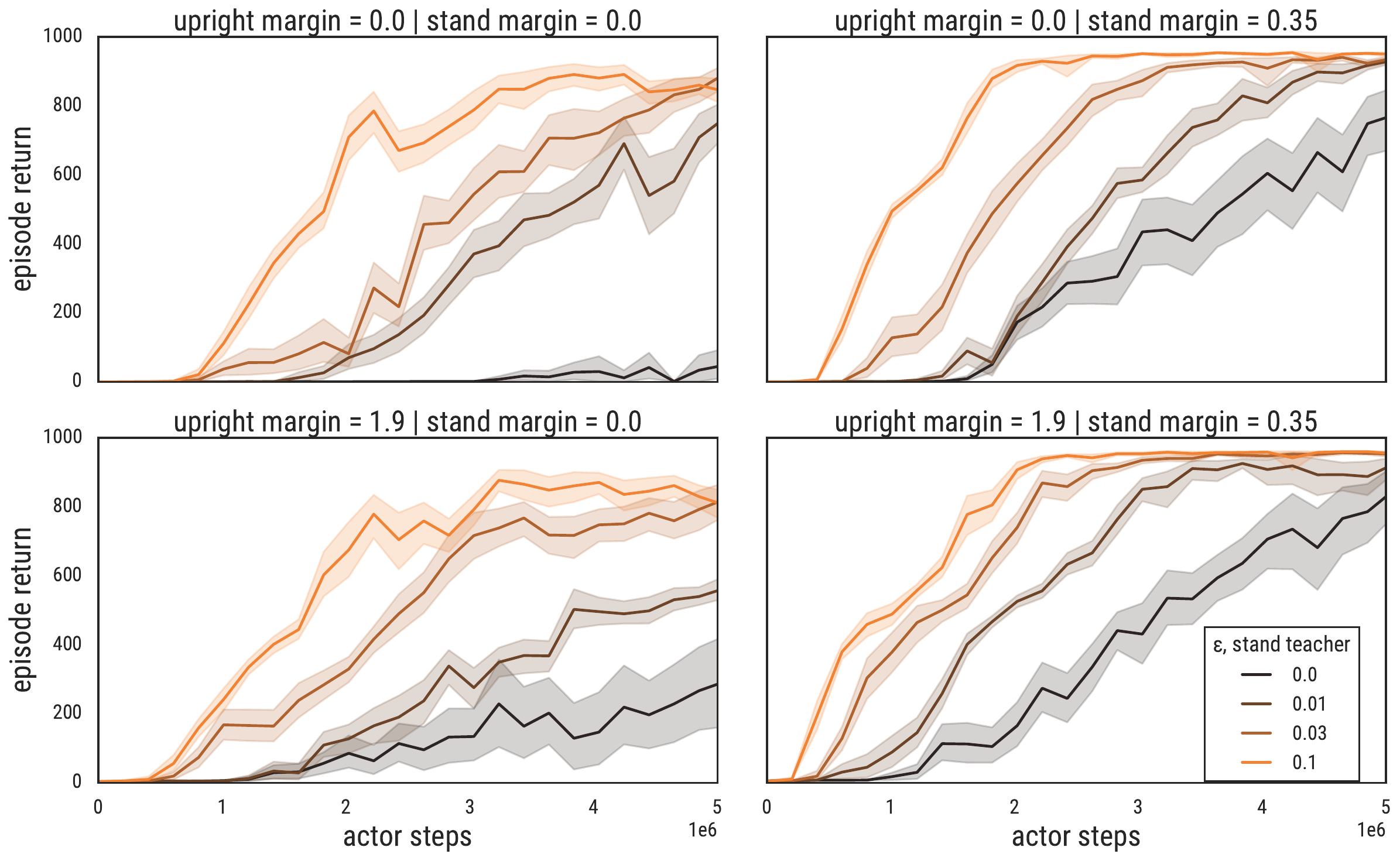}
    \caption{{\bf Learning curves for 10 seeds of the {\tt stand} task with a {\tt stand} teacher}, for different values of $\epsilon_{\mathrm{teacher}}$ (shown in different colors), with $\epsilon_{\mathrm{task}}=0.1$. The different panels correspond to different values of reward sparsity; lower values correspond to more sparsity in the the reward term. The episode return is shown for actor steps at intervals of $1e5$. The thick lines and shading correspond to the mean values and a 95\% confidence interval, respectively.}
    \label{fig:stand-stand}
\end{figure}

\autoref{fig:stand-stand} shows the effect of using the {\tt stand} teacher policy on the {\tt stand} task. In all settings of reward sparsity, we see that increasing the constraint imposed by the teacher, through $\epsilon_{\mathrm{teacher}}$, leads to a speed up of learning. In the sparse reward settings, the agents with teachers do not see much reduction in learning speed versus the agents in the dense reward setting. This is because a shaping reward is effectively provided by the teacher. In contrast, the agents without teachers learn more slowly. 

\begin{figure}[h!]
    \center
    \includegraphics[width=1\columnwidth]
    {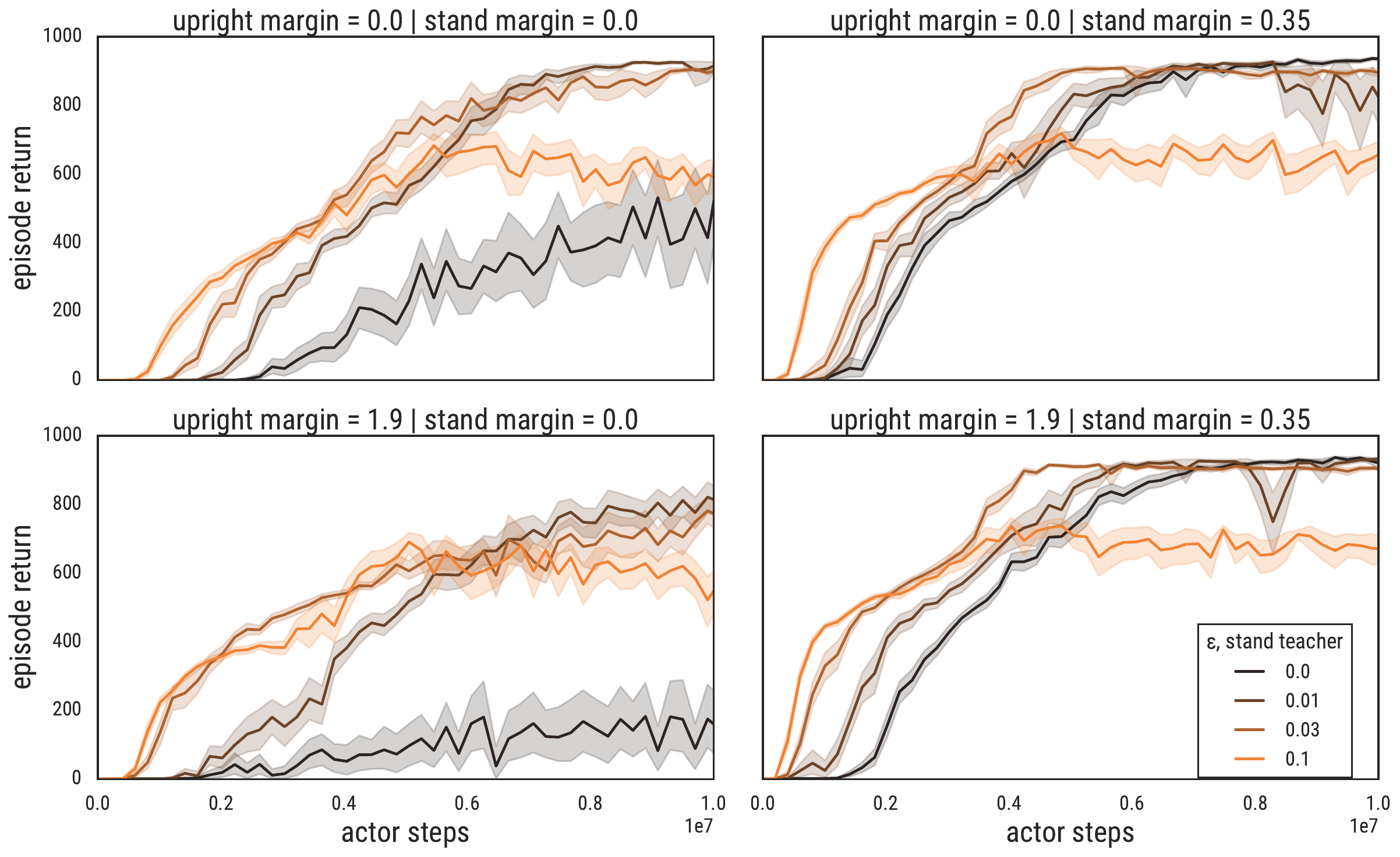}
    \caption{{\bf Learning curves for 10 seeds of the {\tt walk} task with a {\tt stand} teacher}, for different values of $\epsilon_{\mathrm{teacher}}$, with $\epsilon_{\mathrm{task}}=0.1$. The notation is the same as that of \autoref{fig:stand-stand}.}
    \label{fig:walk-stand}
\end{figure}
\autoref{fig:walk-stand} shows the effect of using the {\tt stand} teacher policy on the {\tt walk} task. In all cases, the agents with teachers learn faster when they are early in learning. Later in learning, there may be a crossover, where some of the agents without teachers start to achieve higher task performance. This is because the agents with teachers have their performance limited by the requirement to adhere to the policy of {\tt stand} teacher, which does not have any noticeable forward velocity. 
The existence and location of a crossover depends on the value of $\epsilon_{\mathrm{teacher}}$.

\subsection{Humanoid domain, using stand and walk teachers}
\label{sec:comp-exp-hum-composedteachers}

In this section, we consider the {\tt humanoid} {\tt walk} task with two teachers, to explore composition with multiple teacher policies. In addition to a {\tt stand} teacher policy, we have a {\tt walk} teacher policy. This policy has been trained on a {\tt humanoid} {\tt walk} task with a modification: the agent is always initialized in an upright orientation.
If the height of its head above the ground is less than a threshold, the episode is terminated. This means that the policy learns how to execute a {\tt walk} from an initial upright orientation. However, it never learns to regain balance. This is to separate the expertise of the {\tt walk} teacher from the {\tt stand} teacher; if the {\tt walk} teacher were able to regain balance, there would be no need for a distinct {\tt stand} teacher. This configuration of teachers means that the teacher policies are relevant to distinct parts of the observation space, and the agent is required to compose policies temporally. 

As in \autoref{sec:comp-exp-human-std}, we consider tasks with and without sparse rewards. In \autoref{sec:comp-exp-human-std}, we already saw the effect of the {\tt stand} teacher in speeding up learning for different values of reward sparsity for the components of the reward corresponding to an upright orientation and head height above a threshold. As such, in this section, we consider sparsity of the reward only for the speed of the {\tt humanoid}. This is specified through the {\tt walk\_margin} parameter.

Noting that the {\tt stand} and {\tt walk} teacher policies are relevant for different parts of the observation space, we encode this information explicitly in the teacher policy. Specifically, we now have only one teacher policy which is a function of the observation, $\pi_\mathrm{teacher}(o)$. When the head height of the {\tt humanoid} is below a threshold, $\pi_\mathrm{teacher}(o)=\pi_\mathrm{teacher,\ stand}$. When the head height of the {\tt humanoid} is above a threshold, $\pi_\mathrm{teacher}(o)=\pi_\mathrm{teacher,\ walk}$. This setting is designed to explore the case of an agent being able to usefully compose between multiple teachers with different specializations relevant to different parts of the observation space, i.e., compose policies temporally.  

\begin{figure}[h!]
    \center
    \includegraphics[width=1\columnwidth]
    {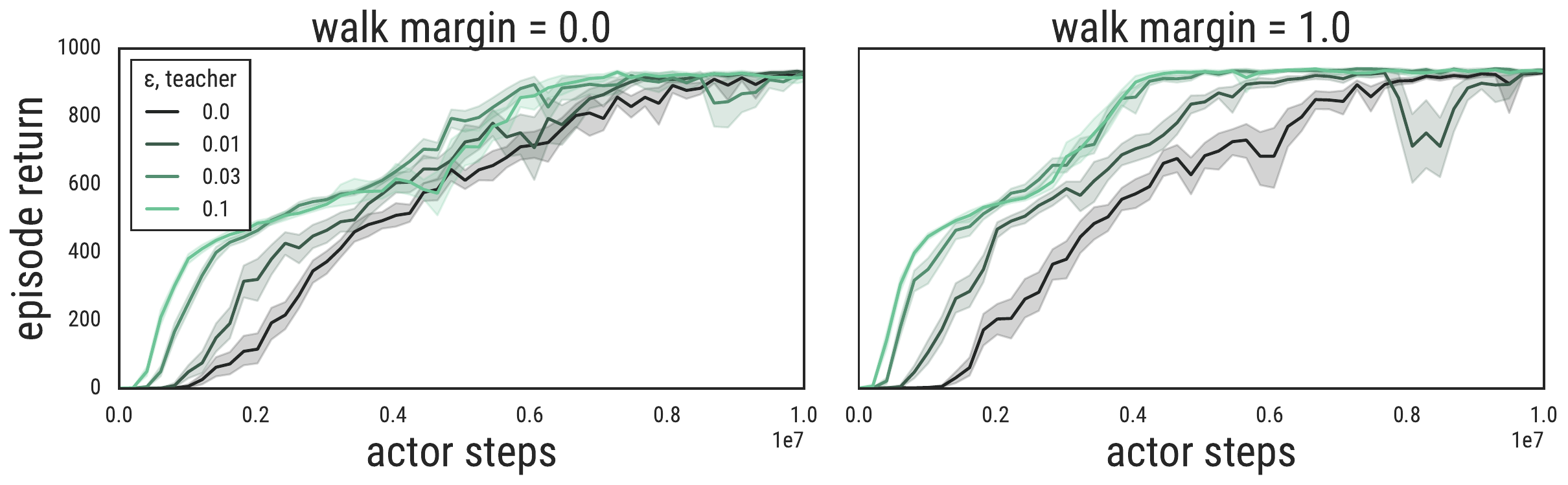}
    \caption{{\bf Learning curves for 10 seeds of the {\tt walk} task with the {\tt stand} and {\tt walk} teachers being active in distinct parts of the observation space.} The episode return is shown for actor steps at intervals of $1e5$. The thick lines and shading correspond to the mean values and a 95\% confidence interval, respectively.}
    \label{fig:walk-mix}
\end{figure}
\autoref{fig:walk-mix} shows learning curves for the {\tt walk} task with the {\tt stand} and {\tt walk} teachers being active in distinct parts of the observation space. In this case, there is a single value of $\epsilon_{\mathrm{teacher}}$ that applies to the active teacher. We see a speed up in learning for the sparse as well as regular settings of the reward. The learning is stable across the different values of $\epsilon_{\mathrm{teacher}}$. Moreover, agents are able to successfully stitch together between the discontinuous sub-policies of the teacher, and likely use the objective corresponding to the task reward to enable this. This offers a flexibility beyond the use of teacher policies as components to a weighted task policy. 

\subsection{Point mass domain} 
\begin{figure}[h!]
    \center
    \includegraphics[width=0.9\columnwidth]{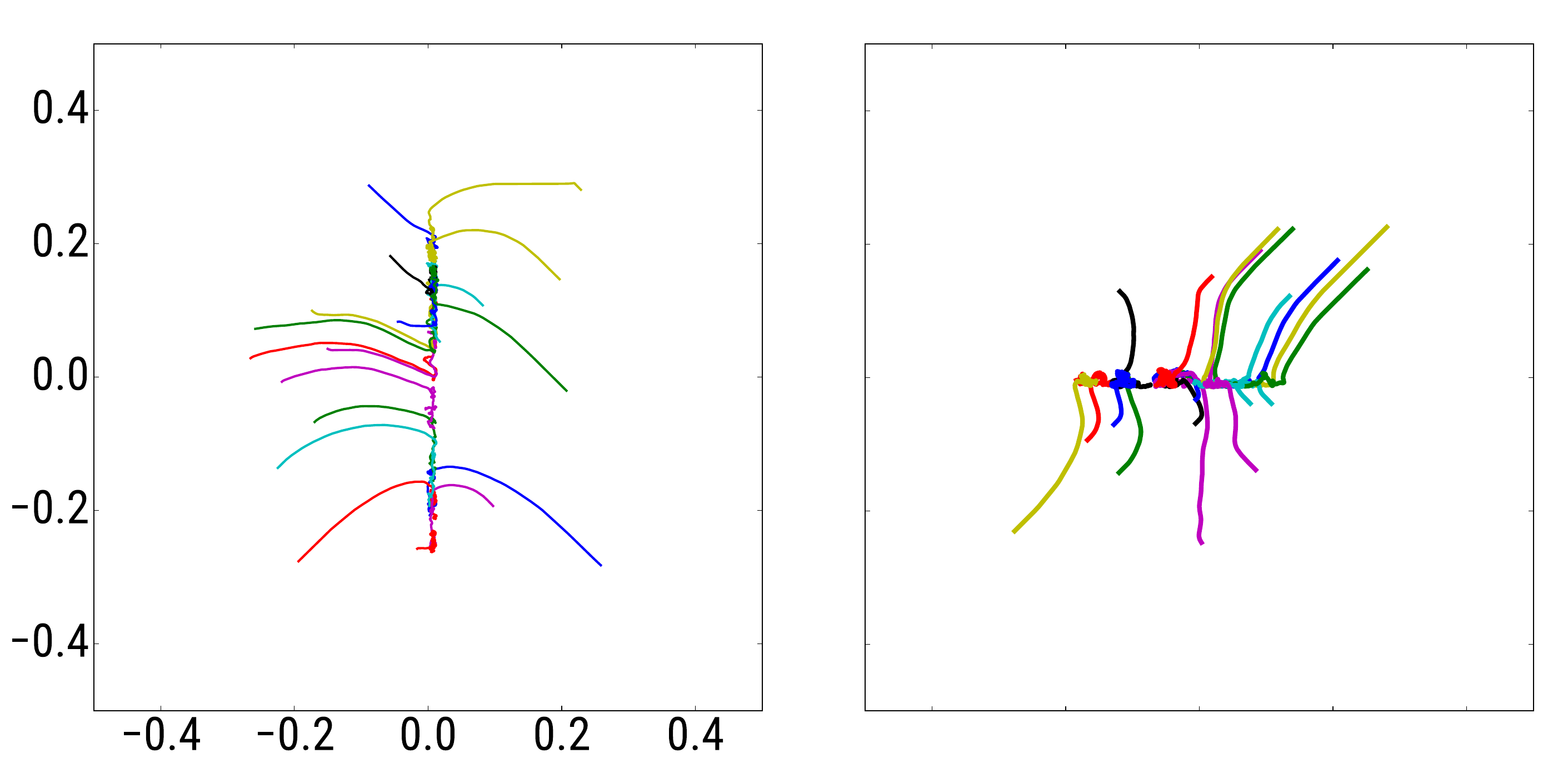}
    \caption{{\bf Trajectories in two-dimensional space for the two trained teacher policies in the point mass domain}, with target $x=0$ (left) and $y=0$ (right), respectively}
    \label{fig:comp-pm-bothteachers}
\end{figure}

We constructed a task where the teacher policies are relevant in the same part of the observation space, i.e., composition of teacher policies must be done spatially. Here, each of the teachers solves a task to go to one of two lines, $x=0$ or $y=0$. \autoref{fig:comp-pm-bothteachers} shows the trajectories of the {\tt point\_mass} in its two-dimensional arena for the policies of the two teachers. 

\begin{figure}[h!]
    \center
    \includegraphics[width=1\columnwidth]{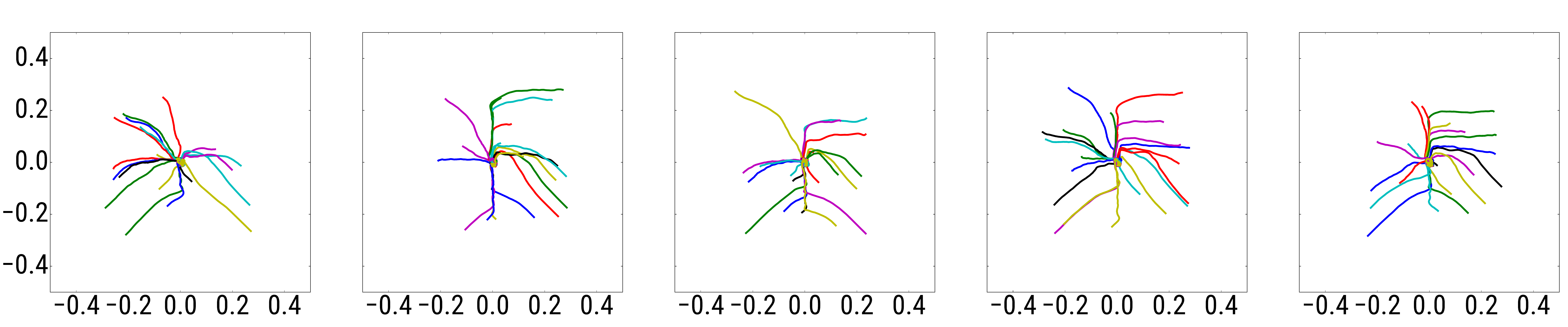}
    \caption{{\bf Trajectories for the composed policy}, with both teachers having equal values of $\epsilon$, and very small $\epsilon_{\mathrm{task}}$. Each panel is a different seed.
    {In all cases the results show the point mass moving towards $x=y=0$.}}
    \label{fig:comp-pm-composed-traj}
\end{figure}
\begin{figure}[h!]
    \center
    \includegraphics[width=0.9\columnwidth]{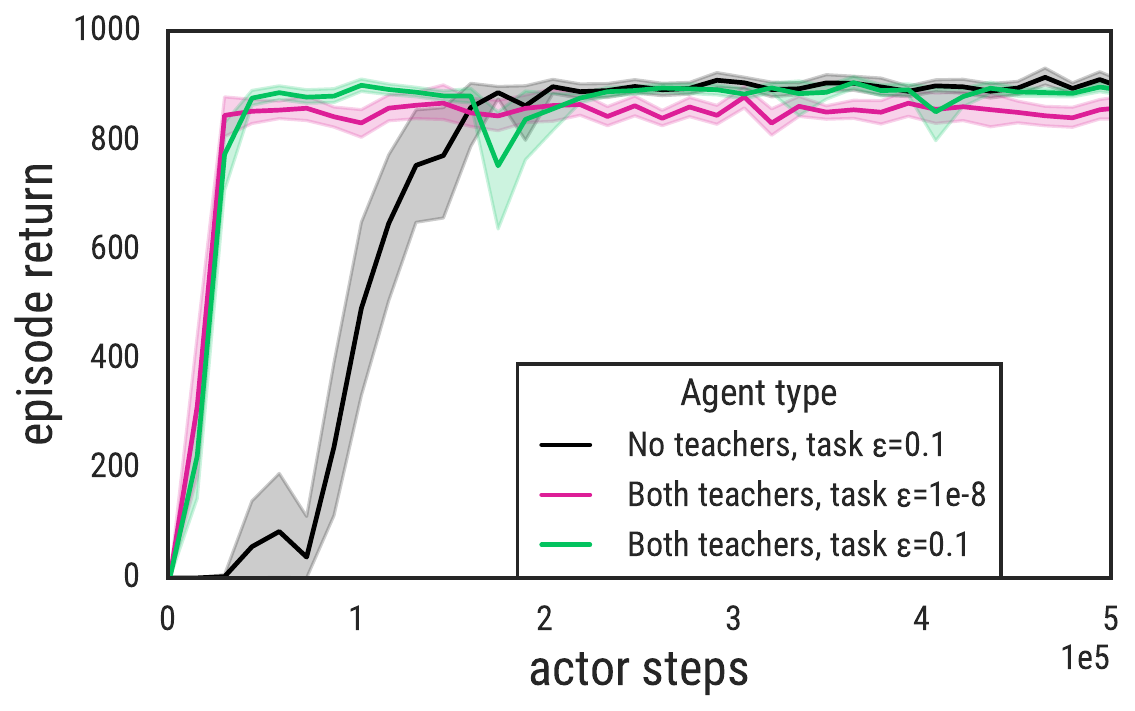}
    \caption{{\bf Learning curves for 5 seeds of the {\tt point\_mass} task with the target at $(0,0)$.}  The black curve corresponds to learning without teachers. The pink and green curves correspond to agents trained with two teachers that go to one of two intersecting lines, $x=0$ or $y=0$; both teachers are active simultaneously, with $\epsilon_{\mathrm{teacher}}=0.1$.
    }
    \label{fig:pointmass-learning}
\end{figure}

\autoref{fig:comp-pm-composed-traj} shows the trajectories of the composed policies in the {\tt point\_mass} domain. We can see that agent policies are successfully composed from the two teachers, without an explicit task reward. For each teacher policy, the {\tt point\_mass} goes to one of two intersecting lines. In the trained policies from a composition of these teachers, the {\tt point\_mass} goes to the intersection of these lines. \autoref{fig:pointmass-learning} shows that the teachers act to speed up learning. 

\subsection{Humanoid domain, agents choosing teachers}
\label{sec:comp-exp-hum-chooseteachers}
The experiments thus far explored selections of teachers where the teacher policies are relevant to overlapping or distinct parts of the observation space. In both of these cases, the relevance of the teacher has been specified through hard-coding the $\epsilon_{\mathrm{teacher},i}$. In this section, we consider an agent that controls the relative preference given to teacher policies, towards learning a successful policy for the task. The method for selecting teachers is described in \autoref{sec:comp-method-comptypes}. We consider the {\tt walk} task in the {\tt humanoid} domain. The teachers are trained in the same way as those of \autoref{sec:comp-exp-hum-composedteachers}.

\begin{figure}
    \centering
    \includegraphics[width=0.9\columnwidth]{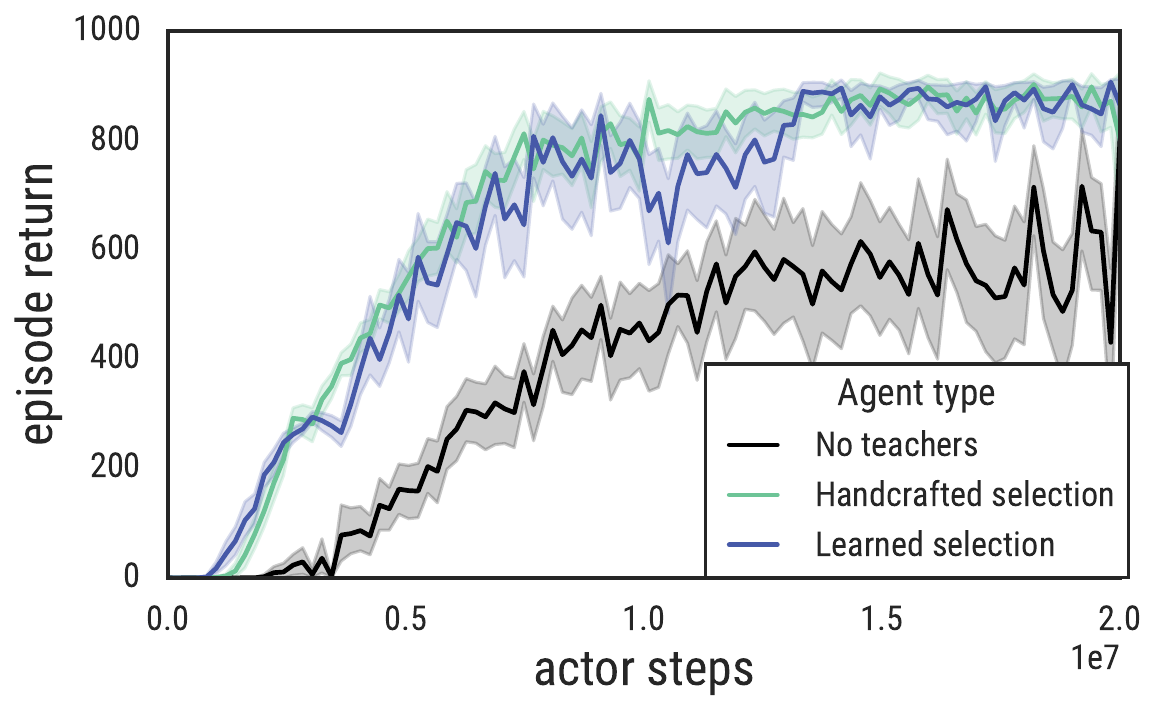}
    \caption{\textbf{Learning curves for 10 seeds of an agent choosing values of $\epsilon_{\mathrm{teacher},i}(o)$}, shown in blue. The baseline comparisons are shown in green and black for handcrafted specification of $\epsilon_{\mathrm{teacher},i}(o)$ and no use of teachers, respectively. Example behavior policies for each of these conditions is indicated by \href{https://www.tinyurl.com/anonrl}{videos}.}
    \label{fig:comp-humanoid-learn-kl-div}
\end{figure}

The blue curve in \autoref{fig:comp-humanoid-learn-kl-div} shows the learning curve for an agent with the ability to choose the preference of teachers on the {\tt humanoid walk} task.
The sparse reward setting of the same task with no teachers is shown by the black curve, and the green curve shows the setting with a handcrafted selection of $\epsilon_{\mathrm{teacher},i}(o)$, equivalent to the experiments in \autoref{sec:comp-exp-hum-composedteachers}, with sparse rewards for the upright, stand and walk components of the reward, and $\epsilon_{\mathrm{teacher}}$ chosen to be the best performance of the user-specified teachers. We see that the purple curve shows an improvement over the baseline, shown in black. As such, the improvement in return compensates for the additional learning cost for searching the space of $\epsilon_{\mathrm{teacher},i}(o)$.

\subsubsection{Characterization of composed policies}
Policies for RL tasks are usually characterized by the reward achieved on the specified task. In composing policies, we may also evaluate the degree to which an agent adheres to specific teacher policies. 
We qualitatively examine the agent policies in comparison to the teacher policies through \href{https://www.tinyurl.com/anonrl}{videos}
of the learned policies. On the {\tt{humanoid walk}} task, agents trained with access to the teacher policies, both in the case of user-specified and agent-selected values of ${\epsilon}_{\mathrm{teacher},i}$ bear a closer resemblance to the teacher policies than agents trained without access to the teacher policies. 

We note that $\epsilon_{\mathrm{teacher}, i}$ may not be an appropriate way to evaluate adherence to teacher policies. This is because a high value of $\epsilon_{\mathrm{teacher}, i}$ requires an agent to adhere to the corresponding policy $\pi_{\mathrm{teacher}, i}$. However, a low value of $\epsilon_{\mathrm{teacher}, i}$ does not imply a low adherence to the teacher policy. Particularly when agents are able to choose $\epsilon_{\mathrm{teacher}, i}$, its value can reduce over the course of learning while the behavior policy persists. 

\section{Discussion and future work}
\label{sec:comp-disc}
We used the MO-MPO algorithm \cite{abdolmaleki2020distributional} to incorporate, in addition to the task objective, a penalty on the KL divergence from teacher policies that are known to be experts for sub-tasks that are characterized either based on the observation (e.g., the {\tt stand} and {\tt walk} tasks in the {\tt humanoid} domain) or based on the action (e.g., move vertically and horizontally in the {\tt point\_mass} domain). In doing so, we successfully show that our agents can compose policies in multiple ways, including both concurrent and sequential composition.
We also observe that, as expected, the behavior of agents with composed policies is closer to the behavior of the teachers than the case where teachers are not incorporated in the learning algorithm.

\subsection{Flexibility in composition} Our method for agents being able to control the influence of teachers through a multi-objective approach, described in \autoref{sec:comp-method-comptypes}, offers flexibility in composition compared to work that uses pre-existing policies as primitives or a lower layer to be built upon, e.g. \citet{qureshi2020composing, peng2019mcp}. 
In \autoref{sec:comp-exp}, we first validate the multi-objective approach to policy composition through experiments with handcrafted values of ${\epsilon}_{\mathrm{teacher},i}(o)$, showing improvement in learning and 
task performance for teachers with relevance to the task in a sequential and concurrent manner. 
In the experiments described in \autoref{sec:comp-exp-hum-chooseteachers}, agents are able to flexibly choose the value of 
$\epsilon_{\mathrm{teacher},i}(o)$ to achieve superior task performance. Indicating the flexibility offered by our approach, in experiments with handcrafted and agent-selected values of $\epsilon_{\mathrm{teacher},i}(o)$, RL agents are able to successfully stitch together between discontinuous teacher policies in the {\tt{humanoid}} domain. 

\subsection{Implication for the design of reward functions} In our experiments in \autoref{sec:comp-exp}, the relative improvements in task performance are greater when the shaping reward is sparse. 
We show that access to teachers more than sufficiently compensates for the lack of shaping reward terms in the {\tt{humanoid}} and {\tt{point\_mass}} domains. Conventionally, in the design of shaping rewards, there are cases where shaping rewards create artifacts that promote behavior(s) corresponding to the specified shaping reward(s) while potentially obstructing behavior(s) corresponding to optimal task performance. Providing agents with access to teachers with some aspects of desirable behavior with the objective function corresponding to a task reward means that such undesirable artifacts may be avoided.

\subsection{On the selection of teachers}  
\subsubsection{Policy composition in action-space} 
This type of composition can be useful for some types of tasks. For example, we may consider $\pi_{\mathrm{teacher,\ walk}}$ to be a teacher policy that is able to walk in some arbitrary direction and $\pi_{\mathrm{teacher,\ hold}}$ to be a teacher policy that is able to hold a box. A {\tt carry} task can be constructed where the agent uses the $\pi_{\mathrm{teacher,\ walk}}$ and $\pi_{\mathrm{teacher,\ hold}}$ for different parts of the action space to carry a box from one location to another. In terms of the notation used in this work, we have allowed  $\epsilon_{\mathrm{teacher},i}$ where $\epsilon_{\mathrm{teacher},i} = \epsilon_{\mathrm{teacher},i}(o)$. In general, we may want $\epsilon_{\mathrm{teacher},i}(o,a)$. The $\epsilon_{\mathrm{teacher},i}(o,a)$ should pass through a bottleneck such that the space of $(\mathcal{O},\mathcal{A})$ for the selection of $\epsilon_{\mathrm{teacher},i}$ is restricted. Otherwise, this may degenerate into a restriction of the observation and action spaces, rather than informative sub-policies. 

\subsubsection{Potential applicability in robotics} Our work is applicable to domains such as robotics with a notion of policies that may be unsafe to execute. By specifying a safe set of policies as teacher policies, agents can be created that adhere to at least one of the allowed set of policies at a given instance. Alternatively, negative values of $\epsilon_{\mathrm{teacher}}(o)$ can be specified to disallow certain configurations. 

\subsubsection{Population-based training} 
For the tasks described thus far, population-based training (PBT) can provide a way to select between the different options of teacher mixtures. PBT may have a greater effect on speeding up learning for tasks that can benefit from a higher dimensional space of teachers, where exploration of the appropriate combinations of $\epsilon_{\mathrm{teacher},i}$ is difficult.

\section*{Acknowledgments}
We thank Ravichandra Addanki, Alexandre Galashov, Mariana Cruz, Leonard Hasenclever, Sandy Huang, John Martin, Josh Merel, Diana Rebmann, Bobak Shahriari, Yuval Tassa and others at DeepMind for advice and feedback during this project.

\bibliographystyle{natbib}

\begin{thebibliography}{24}
\providecommand{\natexlab}[1]{#1}

\bibitem[{Abdolmaleki et~al.(2020)Abdolmaleki, Huang, Hasenclever, Neunert,
  Song, Zambelli, Martins, Heess, Hadsell, and
  Riedmiller}]{abdolmaleki2020distributional}
Abdolmaleki, A.; Huang, S.~H.; Hasenclever, L.; Neunert, M.; Song, H.~F.;
  Zambelli, M.; Martins, M.~F.; Heess, N.; Hadsell, R.; and Riedmiller, M.
  2020.
\newblock A Distributional View on Multi-Objective Policy Optimization.
\newblock \emph{arXiv preprint arXiv:2005.07513}.

\bibitem[{Barreto et~al.(2019)Barreto, Borsa, Hou, Comanici, Ayg{\"u}n, Hamel,
  Toyama, Mourad, Silver, and Precup}]{barreto2019option}
Barreto, A.; Borsa, D.; Hou, S.; Comanici, G.; Ayg{\"u}n, E.; Hamel, P.;
  Toyama, D.; Mourad, S.; Silver, D.; and Precup, D. 2019.
\newblock The option keyboard: Combining skills in reinforcement learning.
\newblock In \emph{Advances in Neural Information Processing Systems},
  13052--13062.

\bibitem[{Barreto et~al.(2020)Barreto, Hou, Borsa, Silver, and
  Precup}]{barreto2020fast}
Barreto, A.; Hou, S.; Borsa, D.; Silver, D.; and Precup, D. 2020.
\newblock Fast reinforcement learning with generalized policy updates.
\newblock \emph{Proceedings of the National Academy of Sciences}, 117(48):
  30079--30087.

\bibitem[{Barth-Maron et~al.(2018)Barth-Maron, Hoffman, Budden, Dabney, Horgan,
  {Tirumala}, Muldal, Heess, and Lillicrap}]{barth2018distributed}
Barth-Maron, G.; Hoffman, M.~W.; Budden, D.; Dabney, W.; Horgan, D.;
  {Tirumala}, D.; Muldal, A.; Heess, N.; and Lillicrap, T. 2018.
\newblock Distributed distributional deterministic policy gradients.
\newblock \emph{arXiv preprint arXiv:1804.08617}.

\bibitem[{Colabrese et~al.(2017)Colabrese, Gustavsson, Celani, and
  Biferale}]{colabrese2017flow}
Colabrese, S.; Gustavsson, K.; Celani, A.; and Biferale, L. 2017.
\newblock Flow navigation by smart microswimmers via reinforcement learning.
\newblock \emph{Physical review letters}, 118(15): 158004.

\bibitem[{Giardina and Mahadevan(2021)}]{giardina2021models}
Giardina, F.; and Mahadevan, L. 2021.
\newblock Models of benthic bipedalism.
\newblock \emph{Journal of the Royal Society Interface}, 18(174): 20200701.

\bibitem[{Heess et~al.(2017)Heess, Tirumala, Sriram, Lemmon, Merel, Wayne,
  Tassa, Erez, Wang, Eslami, Riedmiller, and Silver}]{heess2017emergence}
Heess, N.; Tirumala, D.; Sriram, S.; Lemmon, J.; Merel, J.; Wayne, G.; Tassa,
  Y.; Erez, T.; Wang, Z.; Eslami, S.; Riedmiller, M.; and Silver, D. 2017.
\newblock Emergence of locomotion behaviours in rich environments.
\newblock \emph{arXiv preprint arXiv:1707.02286}.

\bibitem[{Hoffman et~al.(2020)Hoffman, Shahriari, Aslanides, Barth-Maron,
  Momchev, Sinopalnikov, Sta\'nczyk, Ramos, Raichuk, Vincent, Hussenot,
  Dadashi, Dulac-Arnold, Orsini, Jacq, Ferret, Vieillard, Ghasemipour, Girgin,
  Pietquin, Behbahani, Norman, Abdolmaleki, Cassirer, Yang, Baumli, Henderson,
  Friesen, Haroun, Novikov, Colmenarejo, Cabi, Gulcehre, Paine, Srinivasan,
  Cowie, Wang, Piot, and de~Freitas}]{hoffman2020acme}
Hoffman, M.~W.; Shahriari, B.; Aslanides, J.; Barth-Maron, G.; Momchev, N.;
  Sinopalnikov, D.; Sta\'nczyk, P.; Ramos, S.; Raichuk, A.; Vincent, D.;
  Hussenot, L.; Dadashi, R.; Dulac-Arnold, G.; Orsini, M.; Jacq, A.; Ferret,
  J.; Vieillard, N.; Ghasemipour, S. K.~S.; Girgin, S.; Pietquin, O.;
  Behbahani, F.; Norman, T.; Abdolmaleki, A.; Cassirer, A.; Yang, F.; Baumli,
  K.; Henderson, S.; Friesen, A.; Haroun, R.; Novikov, A.; Colmenarejo, S.~G.;
  Cabi, S.; Gulcehre, C.; Paine, T.~L.; Srinivasan, S.; Cowie, A.; Wang, Z.;
  Piot, B.; and de~Freitas, N. 2020.
\newblock Acme: A research framework for distributed reinforcement learning.
\newblock Technical report.

\bibitem[{Mishra, van Rees, and Mahadevan(2020)}]{mishra2020coordinated}
Mishra, S.; van Rees, W.~M.; and Mahadevan, L. 2020.
\newblock Coordinated crawling via reinforcement learning.
\newblock \emph{Journal of the Royal Society Interface}, 17(169): 20200198.

\bibitem[{Mnih et~al.(2015)Mnih, Kavukcuoglu, Silver, Rusu, Veness, Bellemare,
  Graves, Riedmiller, Fidjeland, Ostrovski, Petersen, Beattie, Sadik,
  Antonoglou, King, Kumaran, Wierstra, Legg, and Hassabis}]{mnih2015human}
Mnih, V.; Kavukcuoglu, K.; Silver, D.; Rusu, A.~A.; Veness, J.; Bellemare,
  M.~G.; Graves, A.; Riedmiller, M.; Fidjeland, A.~K.; Ostrovski, G.; Petersen,
  S.; Beattie, C.; Sadik, A.; Antonoglou, I.; King, H.; Kumaran, D.; Wierstra,
  D.; Legg, S.; and Hassabis, D. 2015.
\newblock Human-level control through deep reinforcement learning.
\newblock \emph{Nature}, 518(7540): 529--533.

\bibitem[{Novati, Mahadevan, and Koumoutsakos(2019)}]{novati2019controlled}
Novati, G.; Mahadevan, L.; and Koumoutsakos, P. 2019.
\newblock Controlled gliding and perching through deep-reinforcement-learning.
\newblock \emph{Physical Review Fluids}, 4(9): 093902.

\bibitem[{Peng et~al.(2019)Peng, Chang, Zhang, Abbeel, and
  Levine}]{peng2019mcp}
Peng, X.~B.; Chang, M.; Zhang, G.; Abbeel, P.; and Levine, S. 2019.
\newblock {MCP}: Learning composable hierarchical control with multiplicative
  compositional policies.
\newblock In \emph{Advances in Neural Information Processing Systems},
  3686--3697.

\bibitem[{Puterman(2014)}]{puterman2014markov}
Puterman, M.~L. 2014.
\newblock \emph{Markov decision processes: discrete stochastic dynamic
  programming}.
\newblock John Wiley \& Sons.

\bibitem[{Qureshi et~al.(2020)Qureshi, Johnson, Qin, Henderson, Boots, and
  Yip}]{qureshi2020composing}
Qureshi, A.~H.; Johnson, J.~J.; Qin, Y.; Henderson, T.; Boots, B.; and Yip,
  M.~C. 2020.
\newblock Composing task-agnostic policies with deep reinforcement learning.
\newblock In \emph{International Conference on Learning Representations}.

\bibitem[{Reddy et~al.(2016)Reddy, Celani, Sejnowski, and
  Vergassola}]{reddy2016learning}
Reddy, G.; Celani, A.; Sejnowski, T.~J.; and Vergassola, M. 2016.
\newblock Learning to soar in turbulent environments.
\newblock \emph{Proceedings of the National Academy of Sciences}, 113(33):
  E4877--E4884.

\bibitem[{Schulman et~al.(2017)Schulman, Wolski, Dhariwal, Radford, and
  Klimov}]{schulman2017proximal}
Schulman, J.; Wolski, F.; Dhariwal, P.; Radford, A.; and Klimov, O. 2017.
\newblock Proximal policy optimization algorithms.
\newblock \emph{arXiv preprint arXiv:1707.06347}.

\bibitem[{Silver et~al.(2016)Silver, Huang, Maddison, Guez, Sifre, Van
  Den~Driessche, Schrittwieser, Antonoglou, Panneershelvam, Lanctot, Dieleman,
  Grewe, Nham, Kalchbrenner, Sutskever, Lillicrap, Leach, Kavukcuoglu, Graepel,
  and Hassabis}]{silver2016mastering}
Silver, D.; Huang, A.; Maddison, C.~J.; Guez, A.; Sifre, L.; Van Den~Driessche,
  G.; Schrittwieser, J.; Antonoglou, I.; Panneershelvam, V.; Lanctot, M.;
  Dieleman, S.; Grewe, D.; Nham, J.; Kalchbrenner, N.; Sutskever, I.;
  Lillicrap, T.; Leach, M.; Kavukcuoglu, K.; Graepel, T.; and Hassabis, D.
  2016.
\newblock Mastering the game of {Go} with deep neural networks and tree search.
\newblock \emph{Nature}, 529(7587): 484.

\bibitem[{Silver et~al.(2018)Silver, Hubert, Schrittwieser, Antonoglou, Lai,
  Guez, Lanctot, Sifre, Kumaran, Graepel, Lillicrap, Simonyan, and
  Hassabis}]{silver2018general}
Silver, D.; Hubert, T.; Schrittwieser, J.; Antonoglou, I.; Lai, M.; Guez, A.;
  Lanctot, M.; Sifre, L.; Kumaran, D.; Graepel, T.; Lillicrap, T.; Simonyan,
  K.; and Hassabis, D. 2018.
\newblock A general reinforcement learning algorithm that masters chess, shogi,
  and {Go} through self-play.
\newblock \emph{Science}, 362(6419): 1140--1144.

\bibitem[{Silver et~al.(2017)Silver, Schrittwieser, Simonyan, Antonoglou,
  Huang, Guez, Hubert, Baker, Lai, Bolton, Chen, Lillicrap, Hui, Sifre, van~den
  Driessche, Graepel, and Hassabis}]{silver2017mastering}
Silver, D.; Schrittwieser, J.; Simonyan, K.; Antonoglou, I.; Huang, A.; Guez,
  A.; Hubert, T.; Baker, L.; Lai, M.; Bolton, A.; Chen, Y.; Lillicrap, T.; Hui,
  F.; Sifre, L.; van~den Driessche, G.; Graepel, T.; and Hassabis, D. 2017.
\newblock Mastering the game of {Go} without human knowledge.
\newblock \emph{Nature}, 550(7676): 354.

\bibitem[{Tassa et~al.(2018)Tassa, Doron, Muldal, Erez, Li, de~Las~Casas,
  Budden, Abdolmaleki, Merel, Lefrancq, Lillicrap, and
  Riedmiller}]{deepmindcontrolsuite2018}
Tassa, Y.; Doron, Y.; Muldal, A.; Erez, T.; Li, Y.; de~Las~Casas, D.; Budden,
  D.; Abdolmaleki, A.; Merel, J.; Lefrancq, A.; Lillicrap, T.; and Riedmiller,
  M. 2018.
\newblock Deep{Mind} Control Suite.
\newblock Technical report.

\bibitem[{Todorov, Erez, and Tassa(2012)}]{todorov2012mujoco}
Todorov, E.; Erez, T.; and Tassa, Y. 2012.
\newblock {MuJoCo}: A physics engine for model-based control.
\newblock In \emph{2012 IEEE/RSJ International Conference on Intelligent Robots
  and Systems}, 5026--5033. IEEE.

\bibitem[{Verma, Novati, and Koumoutsakos(2018)}]{verma2018efficient}
Verma, S.; Novati, G.; and Koumoutsakos, P. 2018.
\newblock Efficient collective swimming by harnessing vortices through deep
  reinforcement learning.
\newblock \emph{Proceedings of the National Academy of Sciences}, 115(23):
  5849--5854.

\bibitem[{Vinyals et~al.(2019)Vinyals, Babuschkin, Czarnecki, Mathieu, Dudzik,
  Chung, Choi, Powell, Ewalds, Georgiev, Oh, Danihelka, Huang, Sifre, Cai,
  Agapiou, Jaderberg, Vezhnevets, Leblond, Pohlen, Dalibard, Budden, Sulsky,
  Molloy, Paine, Pfaff, Wu, Ring, Yogatama, W{\"u}nsch, McKinney, Smith,
  Schaul, Lillicrap, Kavukcuoglu, , Hassabis, Apps, and
  Silver}]{vinyals2019grandmaster}
Vinyals, O.; Babuschkin, I.; Czarnecki, W.~M.; Mathieu, M.; Dudzik, A.; Chung,
  J.; Choi, D.~H.; Powell, R.; Ewalds, T.; Georgiev, P.; Oh, J.; Danihelka, I.;
  Huang, A.; Sifre, L.; Cai, T.; Agapiou, J.~P.; Jaderberg, M.; Vezhnevets,
  A.~S.; Leblond, R.; Pohlen, T.; Dalibard, V.; Budden, D.; Sulsky, Y.; Molloy,
  J.; Paine, T.~L.; Pfaff, T.; Wu, Y.; Ring, R.; Yogatama, D.; W{\"u}nsch, D.;
  McKinney, K.; Smith, O.; Schaul, T.; Lillicrap, T.; Kavukcuoglu, K.; ;
  Hassabis, D.; Apps, C.; and Silver, D. 2019.
\newblock Grandmaster level in {StarCraft II} using multi-agent reinforcement
  learning.
\newblock \emph{Nature}, 575(7782): 350--354.

\bibitem[{Wurman et~al.(2022)Wurman, Barrett, Kawamoto, MacGlashan,
  Subramanian, Walsh, Capobianco, Devlic, Eckert, Fuchs, Gilpin, Khandelwal,
  Kompella, Lin, MacAlpine, Oller, Seno, Sherstan, Thomure, Aghabozorgi,
  Barrett, Douglas, Whitehead, Dürr, Stone, Spranger, and
  Kitano}]{wurman2022outracing}
Wurman, P.~R.; Barrett, S.; Kawamoto, K.; MacGlashan, J.; Subramanian, K.;
  Walsh, T.~J.; Capobianco, R.; Devlic, A.; Eckert, F.; Fuchs, F.; Gilpin, L.;
  Khandelwal, P.; Kompella, V.; Lin, H.; MacAlpine, P.; Oller, D.; Seno, T.;
  Sherstan, C.; Thomure, M.~D.; Aghabozorgi, H.; Barrett, L.; Douglas, R.;
  Whitehead, D.; Dürr, P.; Stone, P.; Spranger, M.; and Kitano, H. 2022.
\newblock Outracing champion {G}ran {T}urismo drivers with deep reinforcement
  learning.
\newblock \emph{Nature}, 602(7896): 223--228.

\end{thebibliography}

\newpage
\onecolumn
\section*{Data Appendix}
\label{sec:app}
\begin{table}[H]
\caption{Parameter settings for experiments, using the {\tt{acme}} codebase \cite{hoffman2020acme}}
    \center
    \begin{tabular}{ll} 
    \toprule[1pt]
    \bf{Parameter} & \bf{Value(s)} \\ 
    \cmidrule[0.8pt]{1-2}
    \multicolumn{2}{l}{\textbf{\emph{Training parameters}}} \\ \cmidrule{1-2}
    Batch size & $256$\\
    Replay buffer size & $10^6$ \\
    Target network update period & $100$ \\
    Samples per insert & $32$\\ 
    Number of actors & $16$, humanoid domain; $1$, point mass domain\\
    Adam learning rate & $10^{-4}$\\ 
    Adam learning rate for dual variables & $10^{-2}$\\ 
    \cmidrule[0.8pt]{1-2}
    \multicolumn{2}{l}{\textbf{\emph{Policy network}}} \\
    \cmidrule{1-2}
    Layer sizes & $(256, 256, 256)$\\ 
    Layer norm on first layer? &$\tt{yes}$ \\
    $\tanh$ on output of layer norm & $\tt{yes}$
    \\
    Activation after each hidden layer & ELU \\
    Take $\tanh$ of action mean? & $\tt{no}$ \\
    Minimum variance &$10^{-12}$ \\
    \cmidrule{1-2}
    \multicolumn{2}{l}{{\emph{Additional policy network for choosing $\epsilon_{\mathrm{teacher},i}(o)$}}} \\ \cmidrule{1-2}
    Layer sizes & $(32, 16)$\\ 
    Activation after each hidden layer & ReLU \\
    Take $\tanh$ of mean $\epsilon_{\mathrm{teacher},i}$? & $\tt{yes}$ \\ 
    \cmidrule[0.8pt]{1-2}
    \multicolumn{2}{l}{\textbf{\emph{Critic network}}} \\
    \cmidrule{1-2}
    Layer sizes & $(512, 512, 256)$\\
    Layer norm on first layer? &$\tt{yes}$ \\
    $\tanh$ on output of layer norm & $\tt{yes}$ \\
    Activation after each hidden layer & ELU \\
    Number of atoms & $51$ \\
    Discount factor & $0.99$ \\ 
\cmidrule[0.8pt]{1-2}    \multicolumn{2}{l}{\textbf{\emph{Algorithm parameters}}} \\
    \cmidrule{1-2}
    Initial temperature, $\eta$
    & $10$ \\
    Actions sampled per state, $N$ & $20$ \\
    KL constraint on the mean of the Gaussian policy & $5 \times 10^{-3}$ \\ 
    KL constraint on the variance of the Gaussian policy & $5 \times 10^{-6}$ \\ 
        \bottomrule
    \end{tabular}
    \label{tbl:app}
\end{table}

\end{document}